\pdfoutput=1

\documentclass[11pt]{article}

\usepackage[]{ACL2023}

\usepackage{times}
\usepackage{latexsym}
\usepackage{graphicx}
\usepackage{amsfonts}
\usepackage{booktabs}
\usepackage{multirow}
\usepackage{makecell}
\usepackage{amsmath}
\usepackage{subfigure}

\usepackage[T1]{fontenc}

\usepackage[utf8]{inputenc}

\usepackage{microtype}

\usepackage{inconsolata}

\title{Back Attention: Understanding and Enhancing \\ Multi-Hop Reasoning in Large Language Models}

 \author{Zeping Yu \\ University of Manchester          \And
         Yonatan Belinkov \\ Technion - IIT, Israel  \And
         Sophia Ananiadou \\ University of Manchester}

\begin{document}
\maketitle
\begin{abstract}
We investigate how large language models perform latent multi-hop reasoning in prompts like ``Wolfgang Amadeus Mozart's mother's spouse is''. To analyze this process, we introduce logit flow, an interpretability method that traces how logits propagate across layers and positions toward the final prediction. Using logit flow, we identify four distinct stages in single-hop knowledge prediction: (A) entity subject enrichment, (B) entity attribute extraction, (C) relation subject enrichment, and (D) relation attribute extraction. Extending this analysis to multi-hop reasoning, we find that failures often stem from the relation attribute extraction stage, where conflicting logits reduce prediction accuracy. To address this, we propose back attention, a novel mechanism that enables lower layers to leverage higher-layer hidden states from different positions during attention computation. With back attention, a 1-layer transformer achieves the performance of a 2-layer transformer. Applied to four LLMs, back attention improves accuracy on five reasoning datasets, demonstrating its effectiveness in enhancing latent multi-hop reasoning ability.
\end{abstract}

\section{Introduction}
Enhancing the multi-hop reasoning capabilities of large language models (LLMs) has become a central research focus in recent studies \cite{openaio1,qi2024mutual,snell2024scaling,luo2024improve}. A widely used approach, chain-of-thought (COT) reasoning \cite{wei2022chain}, improves accuracy by explicitly articulating intermediate reasoning steps. Many studies have expanded on this idea by generating explicit reasoning chains to further enhance performance \cite{zhou2022least,creswell2022selection,shum2023automatic,yao2024tree}. However, these methods often require substantial computational resources due to multiple inference steps or extensive sampling, leading to high costs and deployment challenges, particularly in large-scale or resource-constrained scenarios.

Therefore, enhancing the ability of latent multi-hop reasoning is crucial for reducing the cost. For example, predicting ``Wolfgang Amadeus Mozart's mother's spouse is'' -> ``Leopold'' demonstrates a model’s ability to internally retrieve and integrate relevant knowledge. Recent studies have investigated the mechanisms underlying latent multi-hop reasoning. Given two hops <e1, r1, e2> and <e2, r2, e3>, where ``e'' represents an ``entity'' and ``r'' a ``relation'', \citet{yang2024large} observe that LLMs can sometimes successfully predict queries like ``The r2 of the r1 of e1 is'' \texttt{->} ``e3'' by latently identifying the bridge entity ``e2''. However, \citet{biran2024hopping} find that the accuracy of latent multi-hop reasoning remains low, even when both individual hops are correct. They hypothesize that the low accuracy arises because factual knowledge is primarily stored in the early layers. If the first hop is resolved too late, the later layers may fail to encode the knowledge for subsequent reasoning steps.

Although latent multi-hop reasoning has been explored, its underlying mechanism remains unclear. First, previous studies primarily focus on the format ``The r2 of the r1 of e1 is''. In this format, the e1 position and the last position inherently obtain the information of r1 and r2, making it unsurprising that information flows between them. A more complex format, ``e1's r1's r2 is'', introduces additional challenges. Due to the autoregressive nature of decoder-only LLMs, earlier positions cannot access later tokens, hindering relational knowledge propagation and leading to lower accuracy than ``The r2 of the r1 of e1 is'' prompts. Second, several studies have shown that the higher attention and feed-forward network (FFN) layers also store knowledge \cite{geva2023dissecting,yu2024neuron}, challenging the prevailing hypothesis about multi-hop reasoning mechanisms. Last, how to leverage interpretability insights to enhance reasoning remains uncertain. Previous studies \cite{sakarvadia2023memory, li2024understanding} rely on model editing methods, which may cause potential risks \cite{gu2024model, gupta2024model}.

\begin{figure}[thb]
  \centering
  \includegraphics[width=0.75\columnwidth]{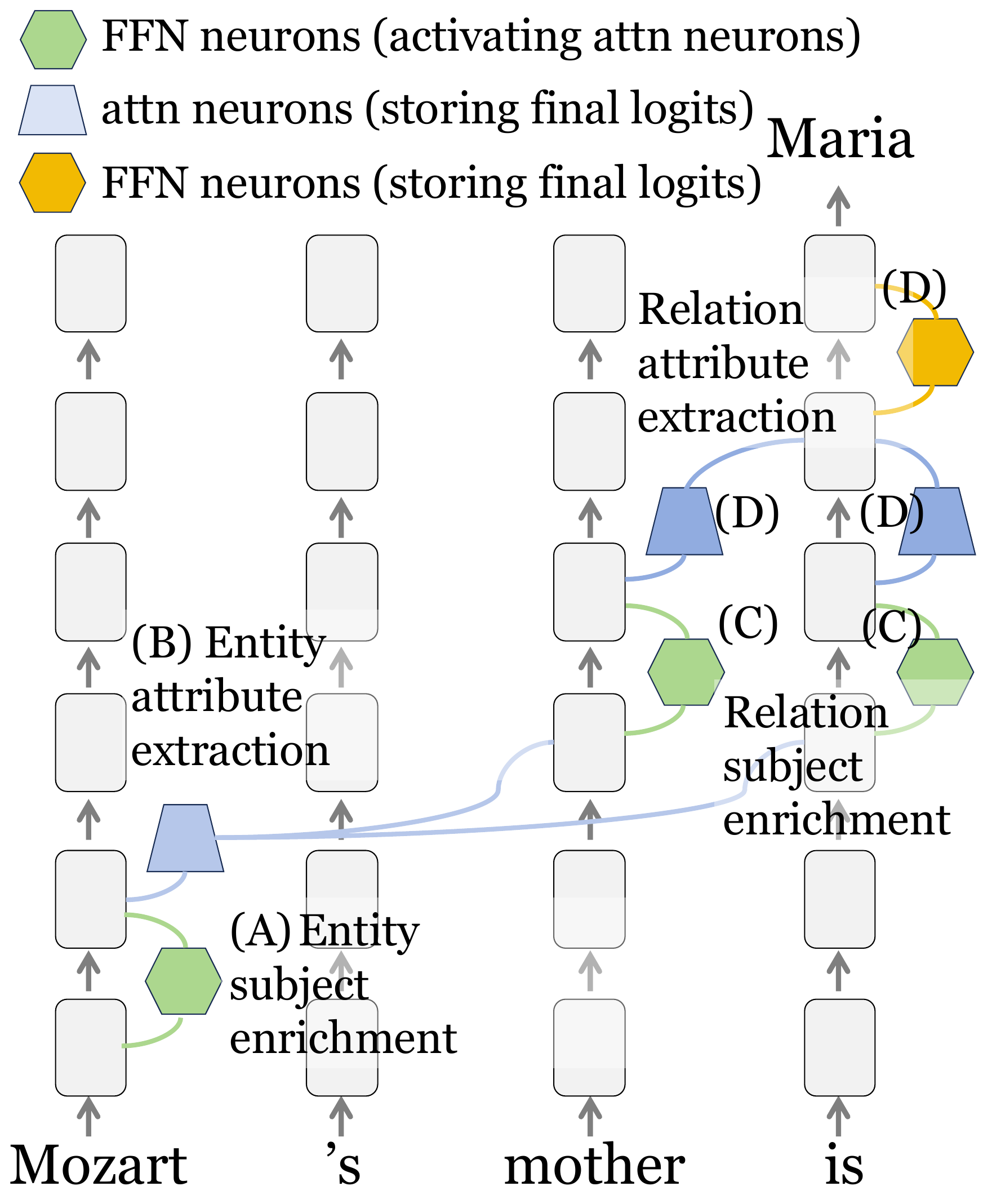}
  \caption{Four stages in single-hop knowledge prediction. At entity position: (A) entity subject enrichment by FFN neurons; (B) entity attribute extraction by attention neurons. At relation and last positions: (C) relation subject enrichments by FFN neurons; (D) relation attribute extraction by attention neurons and FFN neurons.}
\vspace{-10pt}
\end{figure}

In this study, we focus on addressing these challenges. First, we propose an innovative interpretability analysis method named ``logit flow'', which analyzes how logits propagate across different layers and positions toward the final prediction on neuron-level. We use logit flow and activation patching \cite{wang2022interpretability} to analyze the mechanism of single-hop knowledge prediction. We examine prompts such as ``e1's r1 is'' \texttt{->} ``e2'', where e1 represents an entity (e.g. Mozart), r1 represents a relation (e.g. mother), and e2 is the correct answer (e.g. Maria), which is also an entity. We find four main stages, as shown in Figure 1: (A) entity subject enrichment by FFN neurons at e1 position, (B) entity attribute extraction by attention neurons at e1 position, (C) relation subject enrichment by FFN neurons at r1 and last positions, and (D) relation attribute extraction by attention neurons and FFN neurons at r1 and last positions. The first two stages align with \citet{geva2023dissecting}, where entity-related features are enriched and extracted (``e1'' \texttt{->} ``e1 features''). Our analysis further reveals that the last two stages integrate these enriched entity features with the relation, facilitating the prediction of the final token (``e1 features \& r1'' \texttt{->} ``e2'').

Next, we use logit flow and activation patching to analyze correct cases and false cases in two-hop reasoning queries like ``e1's r1's r2 is'', where the correct answer is ``e3'' and the false answer is ``e2''. In false cases, the relation attribute extraction stage strongly captures r1 position's high layer information. Since this attribution occurs at a later stage than when the model encodes ``e2'' -> ``e2 features'' and ``e2 features \& r2'' -> ``e3'', it reinforces e2 more than e3, ultimately reducing two-hop reasoning accuracy. Based on the interpretability findings, we propose an innovative method named ``back attention'' to enhance the multi-hop ability, which allows lower layers to capture higher hidden states. When trained from scratch on arithmetic tasks, a 1-layer transformer with back attention achieves the accuracy of a 2-layer transformer. When applied to four LLMs, back attention boosts accuracy across five reasoning datasets, highlighting its effectiveness in improving multi-hop reasoning ability.

Overall, our contributions are as follow:

a) We introduce logit flow, an innovative interpretability method that traces how logits propagate across layers and positions. We demonstrate its effectiveness in both single-hop and multi-hop reasoning. Specifically, for single-hop knowledge prediction, we identify four key stages: entity subject enrichment, entity attribute extraction, relation subject enrichment, and relation attribute extraction.

b) We apply logit flow to analyze both correct and incorrect multi-hop reasoning cases. Our findings reveal that failures often stem from the relation attribute extraction stage, where conflicting logits disrupt accurate predictions.

c) We propose back attention, a novel technique that enhances feature capture in lower layers by integrating higher-level information. This method is effective both for training from scratch and for adapting pretrained LLMs.

\section{Experimental Settings}
In Section 3 and 4, we use the TwoHop reasoning dataset \cite{biran2024hopping}. Each data instance contains two hops like <e1, r1, e2> and <e2, r2, e3>, where e1, e2, e3 are entities and r1, r2 are relations. For instance, <Wolfgang Amadeus Mozart, mother, Maria Anna Mozart> and <Maria Anna Mozart, spouse, Leopold Mozart> represent two such hops.

We formulate prompts for first-hop, second-hop, and two-hop queries as ``e1's r1 is'', ``e2's r2 is'', and ``e1's r1's r2 is'', respectively. Following \citet{biran2024hopping}, we remove shortcut cases \cite{ju2024investigating} and retain the instances where both the first-hop and two-hop predictions are correct. Then we exclude ⟨e1, e2, e3⟩ triplets appearing fewer than 30 times, ensuring that the model has sufficient exposure to the retained knowledge types. To prevent excessive data duplication, we limit the number of cases where the correct answer e3 appears more than five times. In Section 3, we analyze 889 cases where the first-hop, second-hop, and two-hop queries are all answered correctly. In Section 4, we focus on 568 cases where e1, e2, and e3 are all human entities. This set includes both correct and incorrect two-hop reasoning cases, enabling a broader evaluation of multi-hop reasoning by comparing successful and failed cases.

\section{Mechanism of Single-Hop Prediction}
In Section 3.1, we introduce the background. In Section 3.2, we introduce the proposed interpretability method ``logit flow''. In Section 3.3, we utilize logit flow method and identify the four stages in single-hop knowledge prediction.

\subsection{Background}

\paragraph{Residual Stream.} To better understand how logit flow captures information propagation in decoder-only LLMs, we first introduce the residual stream \cite{elhage2021mathematical}. Given an input sentence $X=[t_1, t_2, ..., t_T]$ with $T$ tokens, the model processes it through residual connections, ultimately producing the probability distribution $y$ over $B$ tokens in vocabulary $V$ for the next token prediction. Each token $t_i$ at position $i$ is transformed into a word embedding $h_i^0 \in \mathbb{R}^{d}$ by the embedding matrix $E \in \mathbb{R}^{B \times d}$. Next, the word embeddings are taken as the $0th$ layer input and transformed by $L+1$ transformer layers ($0th-Lth$). The output of layer $l$ is the sum of the layer input, the attention layer output $A_i^l$ and the FFN layer output $F_i^l$:
\begin{equation}
h_i^l = h_i^{l-1} + A_i^{l} + F_i^{l}
\end{equation}
The probability distribution $y$ is computed by multiplying $h_T^L$ (the final layer $L$ output at the last position $T$) and the unembedding matrix $E_u \in \mathbb{R}^{B \times d}$.
\begin{equation}
y = softmax(E_u \, h_T^{L})
\end{equation}
The attention layer output $A_i^l$ can be regarded as the sum of vectors on different heads and positions:

\begin{equation}
A_i^l = \sum_{j=1}^H \sum_{p=1}^T \alpha_{i,j,p}^l W^o_{j,l} (W^v_{j,l} h_p^{l-1})
\end{equation}
\begin{equation}
\alpha_{i,j,p}^l = softmax(W^q_{j,l} h_i^{l-1} \cdot W^k_{j,l} h_p^{l-1})
\end{equation}
where $H$ is the head number and $\alpha$ is the attention score. $W^q$, $W^k$, $W^v$, $W^o$ are the query, key, value and output matrices in each attention head.

\paragraph{FFN and attention neurons.} 
Based on the computation of FFN output (Eq.5), \citet{geva2020transformer} find that the FFN output is a weighted sum of neurons, where each neuron's contribution is determined by its learned weights and input interactions:
\begin{equation}
F_i^l = W_{fc2}^l\sigma (W_{fc1}^l (h_i^{l-1}+A_i^l))
\end{equation}
\begin{equation}
F_i^l = \sum_{k=1}^N {m_{i,k}^l fc2_{k}^l}
\end{equation}
\begin{equation}
m_{i,k}^l = \sigma (fc1_k^l \cdot (h_i^{l-1}+A_i^l))
\end{equation}
Here, $fc2_k^l$ is the $kth$ column of the second MLP $W_{fc2}^l \in \mathbb{R}^{d \times N}$. Its coefficient score $m$ is computed by the inner product between the residual output and $fc1_k^l$ (the $kth$ row of the first MLP $W_{fc1}^l \in \mathbb{R}^{N \times d}$). Similarly, in attention mechanisms, neuron activations are influenced by key-value transformations \cite{yu2024neuron}. These activations shape how information is stored and propagated through layers, ultimately influencing the model’s predictions:
\begin{equation}
A_i^l = \sum_{j=1}^H \sum_{p=1}^T \sum_{e=1}^{d/H} \alpha_{i,j,p}^l \beta_{j,p,e}^l wo_{j,e}^l
\end{equation}
\begin{equation}
\beta_{j,p,e}^l = wv_{j,e}^l \cdot h_p^{l-1}
\end{equation}
Here, $wo_{j,e}^l$ is the $eth$ column of $W^o_{j,l}$, whose coefficient score $\alpha\beta$ is computed by the inner product between the layer input $h_p^{l-1}$ and $wv_{j,e}^l$ (the $eth$ row of $W^v_{j,l}$), combined with the attention score $\alpha$. 

In this study, we define: 1) A \textbf{subvalue} as the column of the second MLP ($fc2$ in FFN and $wo$ in the attention head). 2) A \textbf{subkey} as the row of the first MLP ($fc1$ in FFN and $wv$ in the attention head). 3) A \textbf{neuron} as the product of the coefficient score and the subvalue (Eq. 6 and Eq. 8).

\subsection{Logit Flow: Tracing the Logits on Different Layers and Positions}

\paragraph{Identifying important neurons in deep layers.} Many studies \cite{dar2022analyzing,geva2022transformer,wang2022interpretability,katz2023visit,yu2024neuron,nikankin2024arithmetic} find that the layer-level and neuron-level vectors in deep layers store logits related to final predictions. When we say a vector stores logits about $s$, we mean that multiplying this vector with the unembedding matrix results in a high log probability for $s$, where the probability of a vector is obtained by multiplying this vector with the unembedding matrix (replacing $h^L_T$ with this vector in Eq.2) \cite{nostalgebraist2020}.

The final vector $h_T^L$ stores large logits about the prediction $s$. The logit increase, $log(p(s|h_T^L))-log(p(s|h_T^0))$, can be decomposed into contributions from $L \times N$ FFN neurons and $L \times H \times T \times d/H$ attention neurons. To identify the neurons in deep layers, we use the log probability increase \cite{yu2024neuron} as importance score:
\begin{equation}
Imp(v^l) = log(p(s|v^l+h^{l-1})) - log(p(s|h^{l-1}))
\end{equation}
If the importance score $Imp(v^l)$ of a neuron $v^l$ is large, it indicates that adding this neuron on its layer input $h^{l-1}$ significantly enhances the log probability of the final prediction $s$. 

\paragraph{Identifying important neurons in shallow layers.} Although shallow neurons typically do not store logits directly related to the final prediction, they can contribute by amplifying the coefficient scores of deeper neurons. For instance, in Eq.9, $\beta$ is computed by the inner product between the attention subkey $wv$ and the layer input $h^{l-1}$, where the layer input is the sum of the neurons from previous layers in the residual stream at this position. 

To analyze this effect, we compute the inner product between the subkey of the 300 most important attention neurons and each preceding FFN neuron, weighting the result by the importance score of the attention neuron. This approach allows us to identify the most influential shallow FFN neurons. If a shallow FFN neuron has a high summed inner product score, it indicates that this neuron activates multiple important attention neurons, thereby indirectly increasing the logits of the final prediction. Unlike previous studies \cite{yu2024neuron}, we retain the inner product of each FFN neuron at every position, rather than summing the scores across all positions. This method enables us to analyze which specific positions and layers contribute the most to activating attention neurons.

\paragraph{Logit flow: an interpretability method for analyzing the logits in different positions and layers.} After identifying the deep FFN and attention neurons that store the final logits, we compute and visualize the sum of their importance scores across different layers and positions. A large score in a specific layer or position indicates that it stores crucial information related to the final prediction. Additionally, we compute and illustrate the weighted sum of inner products of FFN neurons at each layer and position, revealing which layers and positions play a significant role in activating important attention neurons. This approach allows us to distinguish the layers and positions that contribute to predictions both directly and indirectly.

\subsection{Four Stages in Single-Hop Prediction} We utilize logit flow to analyze 889 first-hop queries (``e1's r1 is'' -> ``e2''). We compute the average scores across all cases using LLama2-7B \cite{touvron2023llama2}. If an entity or relation consists of multiple BPE tokens, we sum the scores of these tokens across their respective positions in each layer. The average scores on each layer and position are illustrated in Figure 2. In this and all subsequent logit flow visualizations, the horizontal axis represents the layers, while the vertical axis represents the positions. Darker colors indicate higher logits at a specific position and layer.

\begin{figure}[thb]
  \centering
  \includegraphics[width=0.8\columnwidth]{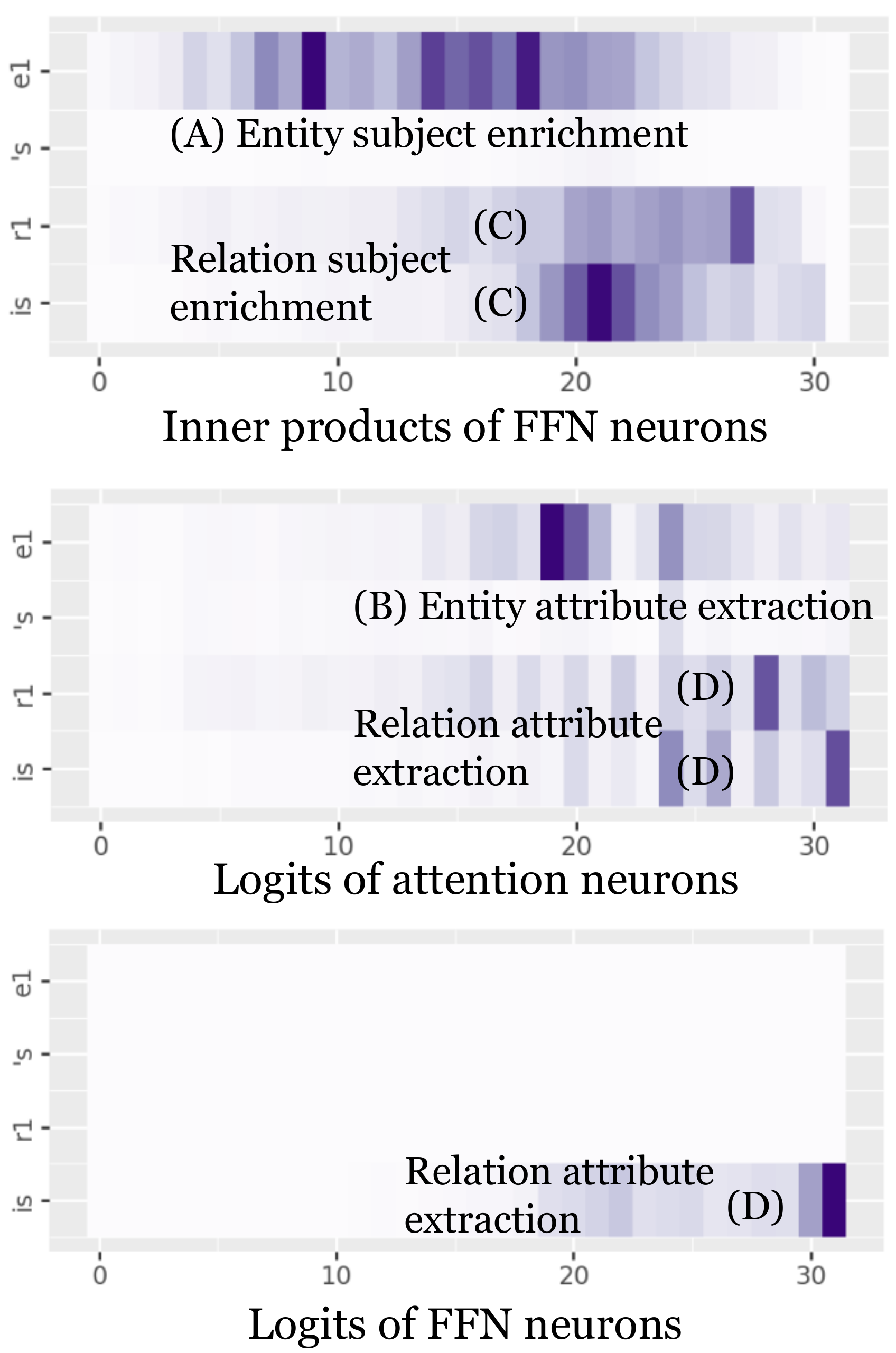}
  \caption{Results of logit flow: ``e1's r1 is'' \texttt{->} ``e2''}
\vspace{-10pt}
\end{figure}

The attention neurons storing logits are distributed across the e1, r1, and last positions, with the layers at e1 being lower than those at r1 and the last position. Similarly, FFN neurons with large inner products are also concentrated at e1, r1, and the last positions, but they generally appear just before the average layers of the attention neurons. The stages at entity position align with the layer-level conclusions in \citet{geva2023dissecting}, where FFN features are activated by the entity's word embeddings and subsequently processed by attention layers.

Additionally, we find that subject enrichment and attribute extraction occur not only at entity position but also at relation and last positions. Due to the autoregressive nature of decoder-only LLMs, the mechanisms at the entity position and r1/last positions differ. At entity position, lower-layer FFN and attention neurons encode knowledge about ``e1 -> e1 features''. In contrast, at the relation and last positions, deeper FFN and attention neurons store knowledge of ``e1 features \& r1 -> e2''. For example, consider ``Mozart's mother is -> Maria'' and ``Mozart's father is -> Leopold''. The hidden states at the position of ``Mozart's'' are identical in both cases, meaning these positions cannot directly determine whether the final prediction is ``Maria'' or ``Leopold''. Instead, at the entity position, lower layers extract Mozart's features containing both ``Maria'' and ``Leopold''. At the relation and last positions, deeper layers refine this information, encoding ``Mozart's features \& mother -> Maria'' and ``Mozart's features \& father -> Leopold'', which enables the model to generate the correct prediction. To verify this, we compute the average logit difference of each layer's hidden state between the correct answer (e.g. Maria) and the conflicting answer (e.g. Leopold) at entity, relation and last positions across all correct human->human cases. The results align with our analysis, detailed in Appendix A. The entity position cannot distinguish the correct answer and the conflicting answer, while the relation and last positions' logit difference start to increase after the entity attribute extraction stage.

We also analyze the logit flow of 889 second-hop cases ``e2's r2 is'' \texttt{->} ``e3'', detailed in Appendix B. Similar to the first-hop results, we observe the same four stages in the second-hop predictions, further validating the single-hop prediction mechanism. In addition, we utilize the activation patching \cite{wang2022interpretability} method to analyze the layer-level information flow, as presented in Appendix C, also observing the importance in entity, relation and last positions. Compared to the layer-level approach, our method provides a neuron-level perspective on information flow, offering a more granular and detailed understanding.

\section{Mechanism of Two-Hop Prediction}
\citet{biran2024hopping} find that the two-hop accuracy remains low, even when both the first-hop and second-hop queries are correct. In this section, we investigate the cause of this phenomenon. We focus on the prompt like ``e1's r1's r2 is'', where the correct answer is ``e3''. We use the logit flow method to analyze the 889 correct two-hop queries, as shown in appendix D. We find that the importance of attention neurons at relation positions is significantly lower than that in single-hop queries. Based on this observation, we hypothesize that the model may incorrectly predict the entity corresponding to ``e1's r1'' or ``e1's r2'' instead of ``e3''. This interference could lead the model to favor intermediate entities over the correct final answer, ultimately reducing the accuracy of two-hop reasoning. 

To verify this, we analyze 568 human->human->human cases with the prompt ``e1's r1's r2 is'' and the correct answer ``e3'' in Llama2-7B, where e1, e2, e3 are all human entities. We compare the ranking of the correct answer ``e3'' against two conflicting answers: ``e1's r1'' and ``e1's r2''. For example, for ``Mozart's mother's spouse is'', the correct answer is ``Leopold'', and the conflicting answers are ``Maria'' (Mozart's mother) and ``Constanze'' (Mozart's spouse). Among 568 cases, 52.3\% correctly predict ``e3'', 42.4\% predict ``e2'' (the answer of ``e1's r1''), and 5.3\% predict the answer of ``e1's r2''. This indicates that the conflicting entities can cause the accuracy decrease. 

\begin{figure}[htb]
  \centering
  \includegraphics[width=0.88\columnwidth]{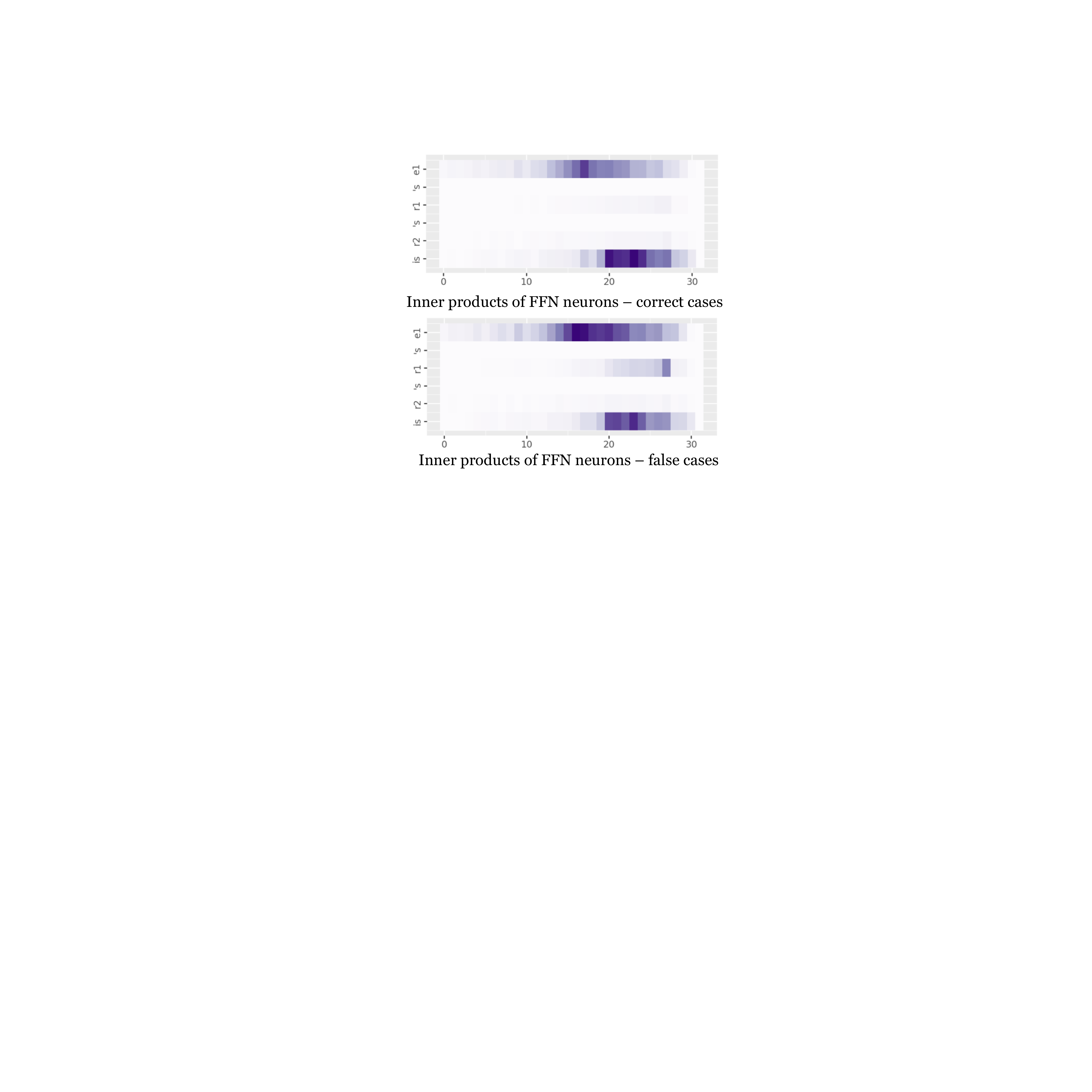}
  \caption{Results of logit flow on correct and false human->human->human cases in Llama2-7B.}
\vspace{-10pt}
\end{figure}

To further investigate this phenomenon, we use the logit flow method to compare correct cases (where the predicted answer is ``e3'') with false cases (where the predicted answer is ``e2''), as shown in Figure 3. We observe that in the false cases, the influence at the r1 position is significantly stronger. The results of activation patching (Appendix E) and Llama3.1-8B \& Llama3.2-3B (Appendix F) reveal a similar trend. This finding appears counterintuitive—why does the model predict the wrong answer when it relies more heavily on the features at the r1 position?

A closer look at the single-hop analysis provides an explanation. In the case of ``e1's r1 is'', the high layers at the r1 position store logits related to ``e2''. Due to the autoregressive nature of decoder-only LLMs, the hidden states at r1 position remain the same in both ``e1's r1 is'' and ``e1's r1's r2 is''. Consequently, when the high-layer information at the r1 position is extracted in ``e1's r1's r2 is'', it inadvertently reinforces the probability of ``e2'', leading to lower accuracy in two-hop reasoning.

This phenomenon can also be understood through the four stages of knowledge storage. In the single-hop analysis (Figure 2), the knowledge of ``e1 -> e1 features'' and ``e2 -> e2 features'' is stored in lower layers (layers 7–20), whereas the knowledge of ``e1 features \& r1 -> e2'' and ``e2 features \& r2 -> e3'' is stored in deeper layers (layers 20–31). In two-hop false cases (Figure 3), when the features at r1 positions, which are related to e2, are extracted at layer 28, they only activate the ``e2 features \& r2 -> e3'' parameters in layers 28–31. Although this process does enhance the probability of e3, it amplifies the probability of e2 even more. This imbalance leads to the model predicting e2 instead of e3, resulting in lower accuracy for two-hop reasoning. From this perspective, our results partially align with the "hopping too late" hypothesis \cite{biran2024hopping}. However, our findings reveal a key difference: while some parameters encoding "e2 \& r2 -> e3" are still activated, their contribution is weaker compared to the direct influence of ``e2''.

\section{Back Attention: Letting Lower Layers Capture Higher-Layer features}
Based on the single-hop mechanism, if we can restore the r1 position's deep layer features back to later positions' shallow layers, the parameters storing ``e2 -> e2 features'' and ``e2 features \& r2 -> e3'' can be activated, thereby strengthening the competitiveness of the correct answer. Motivated by this, we propose an innovative technique, ``back attention'', to allow the lower layers capture higher features. The computations of the original attention output $A$ and the back attention output $B$ are shown in Eq. 11-12. In the original attention computation, the query, key, and value vectors are computed by the hidden states $\mathbf{h}$ on the same layer:
\begin{equation}
\begin{small}
\text{A} = \text{Softmax} \left( \frac{\mathbf{h} \mathbf{W}^q (\mathbf{h} \mathbf{W}^k)^\top}{\sqrt{d'}} \right) (\mathbf{h} \mathbf{W}^v) \mathbf{W}^{o}.
\end{small}
\end{equation}
In contrast, back attention modifies this mechanism by computing queries from a lower source layer $\mathbf{hs}$ while obtaining keys and values from a target layer $\mathbf{ht}$, which are the hidden states on a higher layer or the stack of all higher layers' hidden states. This adjustment allows a lower layer to capture richer representations stored in higher layers:
\begin{equation}
\begin{small}
\text{B} = \text{Softmax} \left( \frac{\mathbf{hs} \mathbf{W}^q_{B} (\mathbf{ht} \mathbf{W}^k_{B})^\top}{\sqrt{d'}} \right) (\mathbf{ht} \mathbf{W}^v_{B}) \mathbf{W}^{o}_{B}.
\end{small}
\end{equation}

\begin{figure}[thb]
  \centering
  \includegraphics[width=0.75\columnwidth]{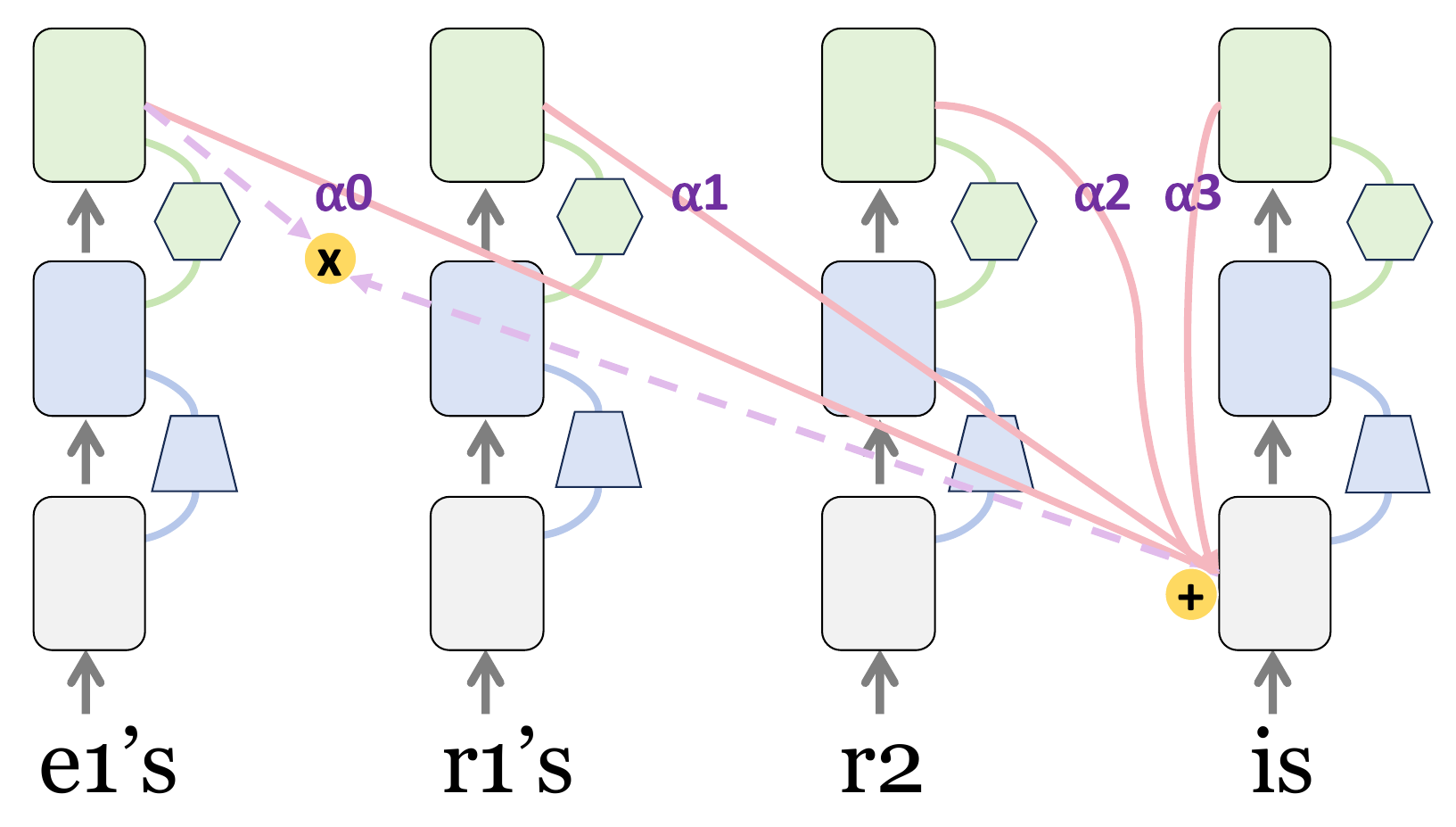}
  \caption{Back attention on a 1-layer transformer.}
\vspace{-10pt}
\end{figure}

Figure 4 illustrates how back attention is integrated into a single-layer transformer. Back attention occurs after the original inference pass, during which the hidden states of all layers and positions are calculated. The query vector is computed from the 0th layer input ($\mathbf{hs}$), while the key and value vectors are computed from the 0th layer output ($\mathbf{ht}$). Then the back attention output $B$ is added back onto the 0th layer input, and recompute the forward pass again. Notably, Figure 4 only shows the back attention computation at the last position, while similar computation happens at all positions on the 0th layer input. Back attention restores high-layer features at different positions using the back attention scores. If the back attention score is 1.0 at r1 position and 0.0 at other positions, it means that the r1 position's 0th layer output is added at the last position's 0th layer input. The results when adding back attention in training and fine-tuning stages are as follows.

\paragraph{Training from scratch: back attention enhances the ability of 1-layer transformer.} We conduct experiments on a 2-digit addition arithmetic dataset. In each training and testing set, there are 12,150 single-sum cases (``a+b=''), and 6,188 double-sum cases (``c+d+e=''), where ``a'', ``b'', ``c'', ``d'', and ``e'' are integers ranging from 0 to 99. The model needs to ``memorize'' the single-sum cases and ``learn'' the double-sum patterns. We utilize the Llama tokenizer, representing each digit as a separate token (e.g., 12 is tokenized as [``1'', ``2'']), ensuring that each token appears sufficiently during training. We compare the results of a 1-layer transformer, a 2-layer transformer, and a 1-layer transformer with back attention. In all models, the dimension is 440 for attention/FFN layers, and 160 for back attention. We use the AdamW optimizer \cite{loshchilov2017decoupled} with a learning rate of 0.0001, a batch size of 64, and a maximum of 500 epochs. 

The accuracy of 1-layer transformer, 1-layer transformer with attention, and 2-layer transformer are 83.8\%, 93.8\%, and 92.5\%, respectively. The details of loss and accuracy are shown in Appendix G. The 2-layer transformer and the 1-layer transformer with back attention converge faster than the 1-layer transformer. Notably, the 1-layer transformer with back attention requires only 56.7\% of the parameters of the 2-layer transformer. Therefore, incorporating back attention during the training stage can significantly enhance the model's performance while reducing parameter requirements.

\begin{figure}[thb]
  \centering
  \includegraphics[width=0.99\columnwidth]{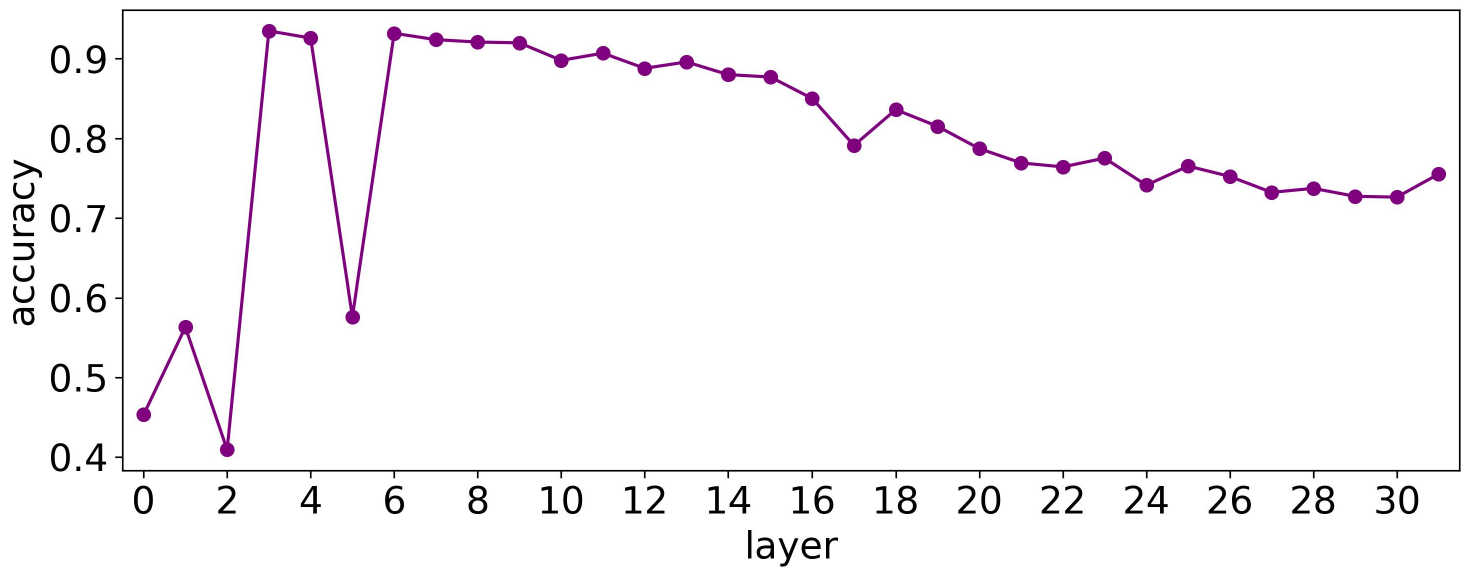}
  \caption{Test accuracy of back attention on each layer.}
\vspace{-10pt}
\end{figure}

\paragraph{Adding back attention in pre-trained LLMs: back attention increases the reasoning accuracy.} Back attention can also be integrated into a pre-trained LLM, using all higher-layer states to compute the keys and values. For instance, if back attention is added to the 6th layer, the keys and values are calculated using the layer outputs of all positions from the 6th layer to the final layer. We add back attention on each layer in Llama-7B \cite{touvron2023llama}, fine-tuning on the double-sum arithmetic cases (the same in the previous section). Figure 5 shows the accuracy when fine-tuning back attention on each layer (freezing LLM parameters), where the original accuracy is 67.1\%. The accuracy across the 0-5 layers exhibits significant fluctuation. Adding back attention to the 6th layer achieves a peak accuracy of 93.2\%, followed by a steady decline compared with higher layers.

\begin{table}[htb]
\centering
\begin{small}
\begin{tabular}{cccccc}
\toprule
  & 1DC & SVAMP & MA & TwoHop & SQA \\
\midrule
Llama3 & 72.7 & 55.7 & 21.1 & 11.5 & 65.1 \\
+backattn & 97.0 & 69.3 & 88.9 & 47.8 & 86.2 \\
\midrule
Llama3.1 & 74.6 & 56.0 & 30.0 & 8.8 & 65.4 \\
+backattn & 98.5 & 70.7 & 86.2 & 42.7 & 87.0 \\
\midrule
Llama3.2 & 49.3 & 44.3 & 15.0 & 6.5 & 62.0 \\
+backattn & 92.9 & 62.0 & 52.8 & 37.0 & 86.3 \\
\midrule
Mistral & 51.9 & 63.0 & 26.1 & 8.8 & 71.5 \\
+backattn & 87.4 & 71.7 & 47.2 & 40.1 & 87.8 \\
\bottomrule
\end{tabular}
\end{small}
\caption{Accuracy (\%) on 5 datasets before/after adding back attention on 6th layer in four LLMs.}
\vspace{-10pt}
\end{table}

Then we do experiments on 5 arithmetic and reasoning datasets 1-Digit-Composite (1DC) \cite{brown2020language}, SVAMP \cite{patel2021nlp}, MultiArith (MA) \cite{roy2016solving}, TwoHop \cite{biran2024hopping}, and StrategyQA (SQA) \cite{geva2021did}. We fine-tune back attention on the 6th layer in Llama3-8B \cite{meta2024introducing}, Llama3.1-8B \cite{dubey2024llama}, Llama3.2-3B \cite{meta2024llama}, and Mistral-7B \cite{jiang2023mistral}. The accuracy is shown in Table 1. On all LLMs, adding back attention achieves a significant accuracy increase. 

\begin{figure}[thb]
  \centering
  \includegraphics[width=0.9\columnwidth]{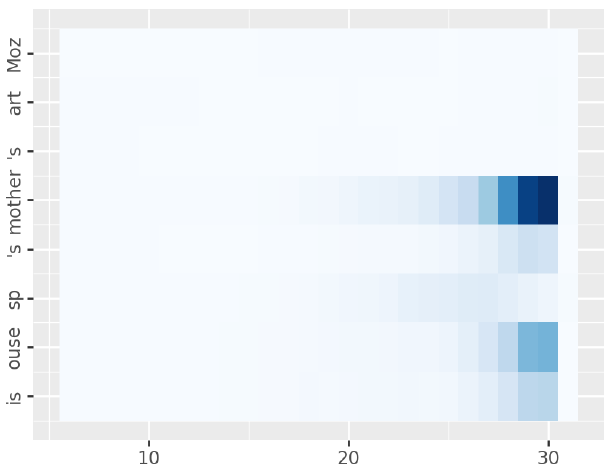}
  \caption{Back attention scores at all positions and higher layers when adding on the 6th layer.}
\vspace{-10pt}
\end{figure}

To evaluate whether back attention functions as intended, we analyze the case ``Mozart's mother's spouse is'' \texttt{->} ``Leopold'' in TwoHop dataset and visualize the back attention scores (darker larger) in Figure 6. Back attention effectively learns to recover ``mother'' position's 27-30 layers' hidden states into the last position's 6th layer. This visualization proves that back attention successfully propagates high-layer information from important positions to lower layers, enabling the model to better utilize knowledge for accurate predictions.

\paragraph{Advantages of back attention.} First, back attention can be incorporated during fine-tuning, enabling flexible enhancement of a powerful pre-trained LLM without retraining the model from scratch. Second, the back attention parameters are remarkably lightweight compared to those of pre-trained LLMs. For instance, the back attention parameters account for just 0.002\% of LLama3's 8 billion parameters. Third, back attention exhibits substantial potential, increasing the average accuracy from 46.9\% to 77.0\% in Llama3.1-8B, even when applied to a single layer. Finally, the visualization of back attention scores (Figure 6) serves as an interpretability tool, offering insights into which positions are most critical for a given task, thereby improving our understanding of the mechanisms.

\section{Related Work}
\subsection{Multi-Hop Reasoning in LLMs}
Improving the reasoning ability of LLMs has become a key focus of recent research \cite{lightman2023let,huang2023large,li2024chain,wang2024chain}. \citet{wei2022chain} use chain-of-thought to enhance the reasoning ability by articulating intermediate steps. \citet{fu2022complexity} propose complexity-based prompting, showing that selecting and generating reasoning chains with higher complexity significantly improves reasoning accuracy. \citet{wang2022self} combine chain-of-thought with the self-consistency decoding strategy, achieving significant improvements by sampling diverse reasoning paths and selecting the most consistent answer. \citet{chen2024self} propose self-play fine-tuning, which enhances LLMs' reasoning abilities by refining their outputs through self-generated data, thereby reducing reliance on human-annotated datasets. \citet{brown2024large} propose scaling inference compute by increasing the number of generated samples, demonstrating significant improvements across tasks like coding and math. \citet{hao2023reasoning,yao2024tree} use tree-based methods to improve the performance.

\subsection{Mechanistic Interpretability}
Mechanistic interpretability \cite{Chris2022} aims to reverse engineer the internal mechanisms of LLMs. Logit lens \cite{nostalgebraist2020} is a widely used method \cite{dar2022analyzing,katz2023visit,yu2024interpreting} to analyze the information of hidden states, by multiplying the vectors with the unembedding matrix. A commonly used localization method is causal mediation analysis \cite{vig2020investigating,meng2022locating,stolfo2023mechanistic,geva2023dissecting}, whose core idea is to compute the change of the output when modifying a hidden state. Another types of studies focus on constructing the circuit in the model \cite{olsson2022context,zhang2023towards,gould2023successor,hanna2024does,wang2022interpretability}. Due to the superposition phenomenon \cite{elhage2022toy,scherlis2022polysemanticity,bricken2023towards}, sparse auto-encoder (SAE) is useful for interpreting the features \cite{gao2024scaling,templeton2024scaling,cunningham2023sparse}. A useful characteristic is the residual stream \cite{elhage2021mathematical}, revealing that the final embedding can be represented as the sum of layer outputs. Furthermore, \citet{geva2020transformer,geva2022transformer} find that the FFN output is the weighted sum of FFN neurons. \citet{yu2024neuron} find that the attention head outputs can also be regarded as the weighted sum of attention neurons. 

While previous neuron-level studies primarily focus on ``localization''—identifying which neurons are important—they often lack a deeper ``analysis'' of how these neurons influence predictions. By applying our logit flow method, we gain a clearer understanding of how neurons are activated and contribute to the final prediction.

\section{Conclusion}
We investigate the mechanisms of latent multi-hop reasoning in LLMs and identify key factors affecting the accuracy. Through our interpretability method logit flow, we uncover four distinct stages in single-hop knowledge prediction: entity subject enrichment, entity attribute extraction, relation subject enrichment, and relation attribute extraction. Analyzing two-hop queries, we find that failures often arise in the relation attribute extraction stage, where conflicting logits lower prediction accuracy. To address this, we propose back attention, a novel method that enables lower layers to access higher-layer hidden states, effectively restoring important features. Back attention significantly enhances reasoning performance, allowing a 1-layer transformer to match the accuracy of a 2-layer transformer. When applied to pre-trained LLMs, it improves accuracy across five datasets and four models, demonstrating its effectiveness in multi-hop reasoning. Overall, our analysis provides new insights and introduces a powerful approach for improving reasoning accuracy in LLMs.

\clearpage
\section{Limitations}
In this study, the interpretability analysis primarily focuses on single-hop and two-hop knowledge queries, which represent specific reasoning scenarios. While these cases provide valuable insights, it is important to acknowledge that other types of reasoning tasks might involve different mechanisms not captured in our analysis. Despite these constraints, the observed performance improvements across a variety of reasoning tasks and LLMs suggest that the proposed back attention method and the derived insights possess a degree of general applicability. Further investigations will be needed to validate these findings on more diverse reasoning tasks and refine the interpretability framework for broader applicability.

In this work, back attention is applied to only a single layer, where it has demonstrated promising results. Nevertheless, back attention can also be extended to two or more layers, potentially yielding even greater improvements. We view the success of the single-layer application as a foundational step, paving the way for future research aimed at exploring and optimizing back attention in more complex and multi-layer configurations.

\bibliography{custom}
\bibliographystyle{acl_natbib}

\clearpage
\appendix

\section{Logit Difference at Different Positions}
\begin{figure}[thb]
  \centering
  \includegraphics[width=0.88\columnwidth]{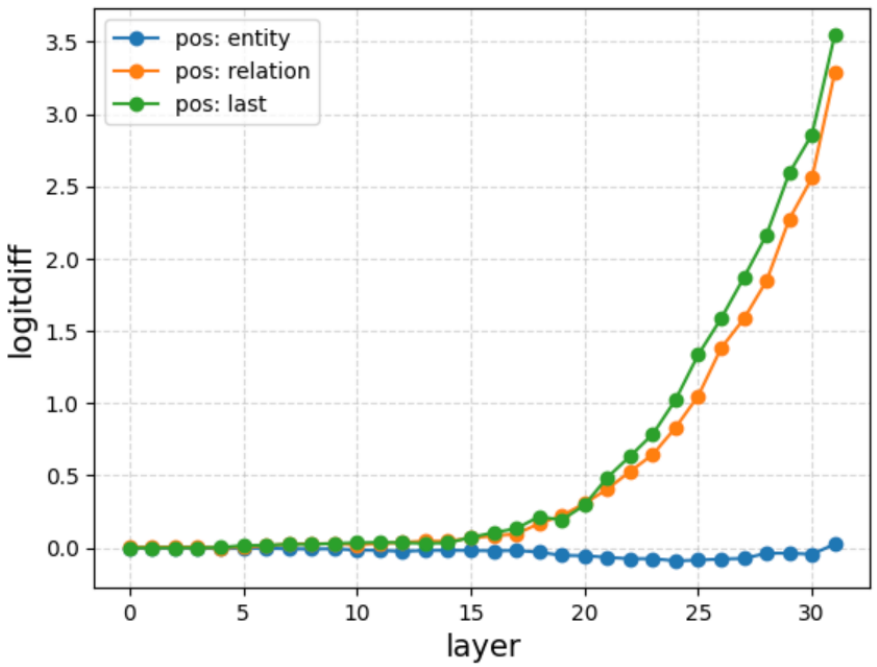}
  \caption{Logit difference at entity, relation and last positions on human->human cases in Llama2-7B. The logit difference is small at entity position, but large on relation and last positions' deep layers.}
\vspace{-10pt}
\end{figure}

We compute the average logit difference at entity, relation and last positions across all correct human -> human cases, shown in Figure 7. Take ``Mozart's mother is -> Maria'' as an example. We compute the logit difference between ``Maria'' and ``Leopold'' (Mozart's father). At the entity position, the logit difference is small on all layers. At the relation and last positions, the logit difference increases sharply after the entity subject enrichment and entity attribute extraction stages (layers 19–20). This indicates that the entity position primarily extracts general features of ``Mozart'', including information relevant to both ``Maria'' and ``Leopold''. In contrast, the deeper layers at the relation and last positions encode specific knowledge, such as ``Mozart's features \& mother -> Maria'' and ``Mozart's features \& father -> Leopold'', which ultimately differentiate the correct prediction.

\section{Results of Logit Flow on Second-Hop Queries in Llama2-7B}
\begin{figure}[thb]
  \centering
  \includegraphics[width=0.8\columnwidth]{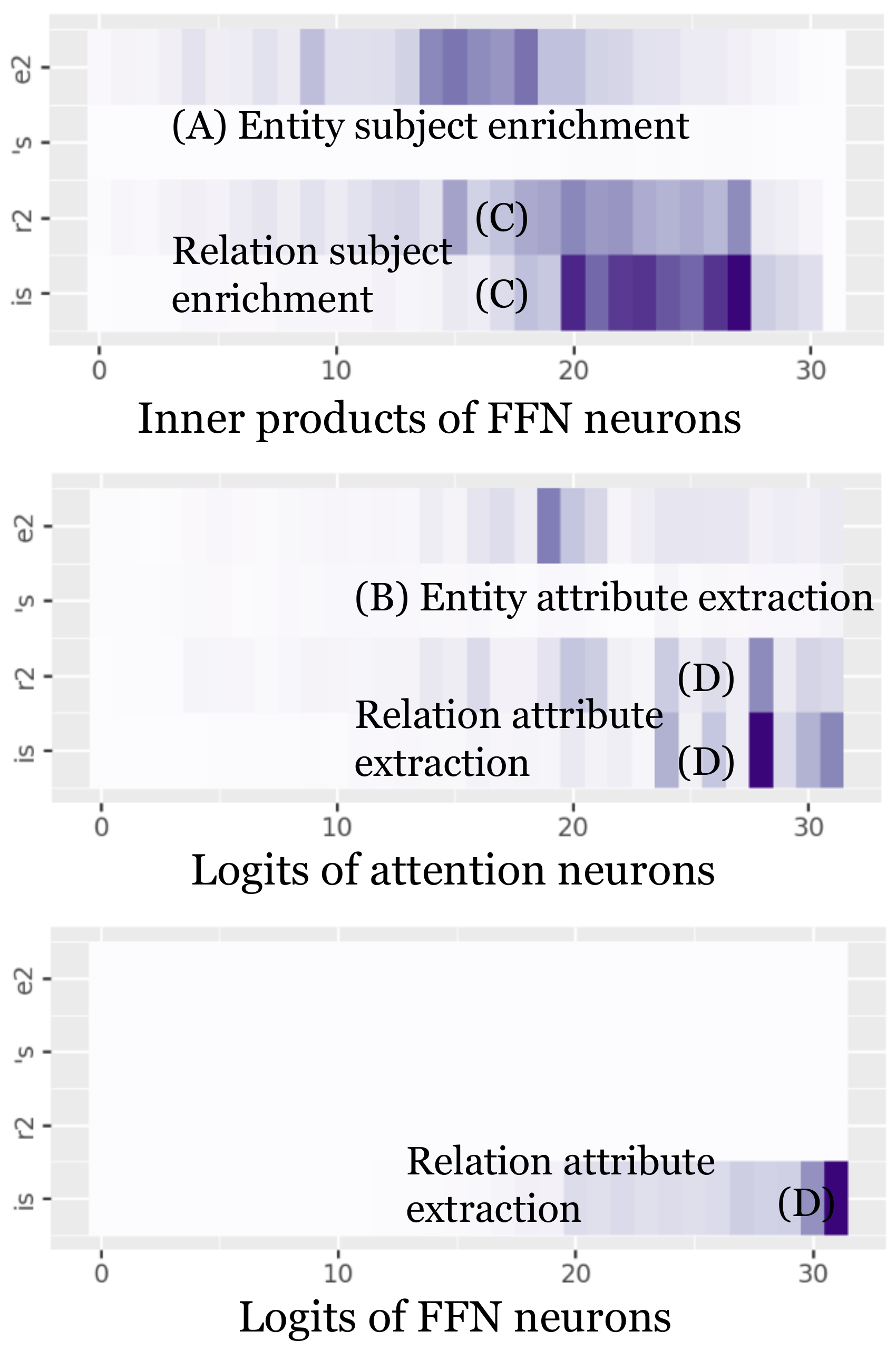}
  \caption{Results of logit flow on second-hop queries ``e2's r2 is'' \texttt{->} ``e3'' in Llama2-7B. There are four similar stages with the first-hop queries: (A) entity subject enrichment, (B) entity attribute extraction, (C) relation subject enrichment, and (D) relation subject extraction.}
\vspace{-10pt}
\end{figure}

The results of logit flow on second-hop queries ``e2's r2 is'' \texttt{->} ``e3'' are shown in Figure 8. There are also four stages existing in the second-hop queries, similar to those in the first-hop queries (Figure 2).

\section{Results of Activation Patching on Single-Hop Queries in Llama2-7B}
The results of activation patching on single-hop queries are shown in Figure 9, using the pyvene \cite{wu2024pyvene} and NNsight \cite{fiotto2024nnsight} libraries. Compared to the logit flow results (Figure 2), the entity and last positions exhibit higher importance, while the relation position appears less significant. This difference arises because activation patching aggregates the importance of both FFN and attention modules into a single visualization. In contrast, the logit flow method distinguishes and separately visualizes the importance of FFN and attention neurons, offering a more granular, neuron-level understanding of the information flow.

\begin{figure}[thb]
  \centering
  \includegraphics[width=0.88\columnwidth]{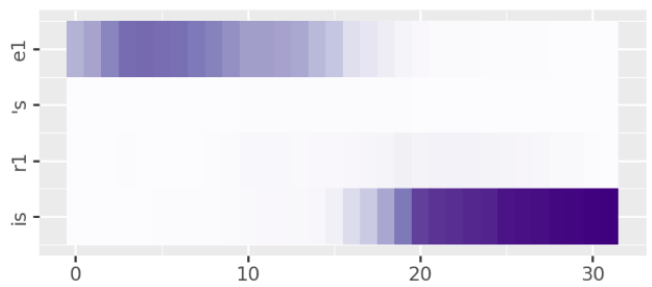}
  \caption{Results of activation patching on single-hop queries in Llama2-7B. Similar to logit flow (but not as obvious as logit flow), there is also importance on r1 position's high layers.}
\vspace{-10pt}
\end{figure}

\section{Results of Logit Flow on Two-Hop Queries in Llama2-7B}
\begin{figure}[thb]
  \centering
  \includegraphics[width=0.8\columnwidth]{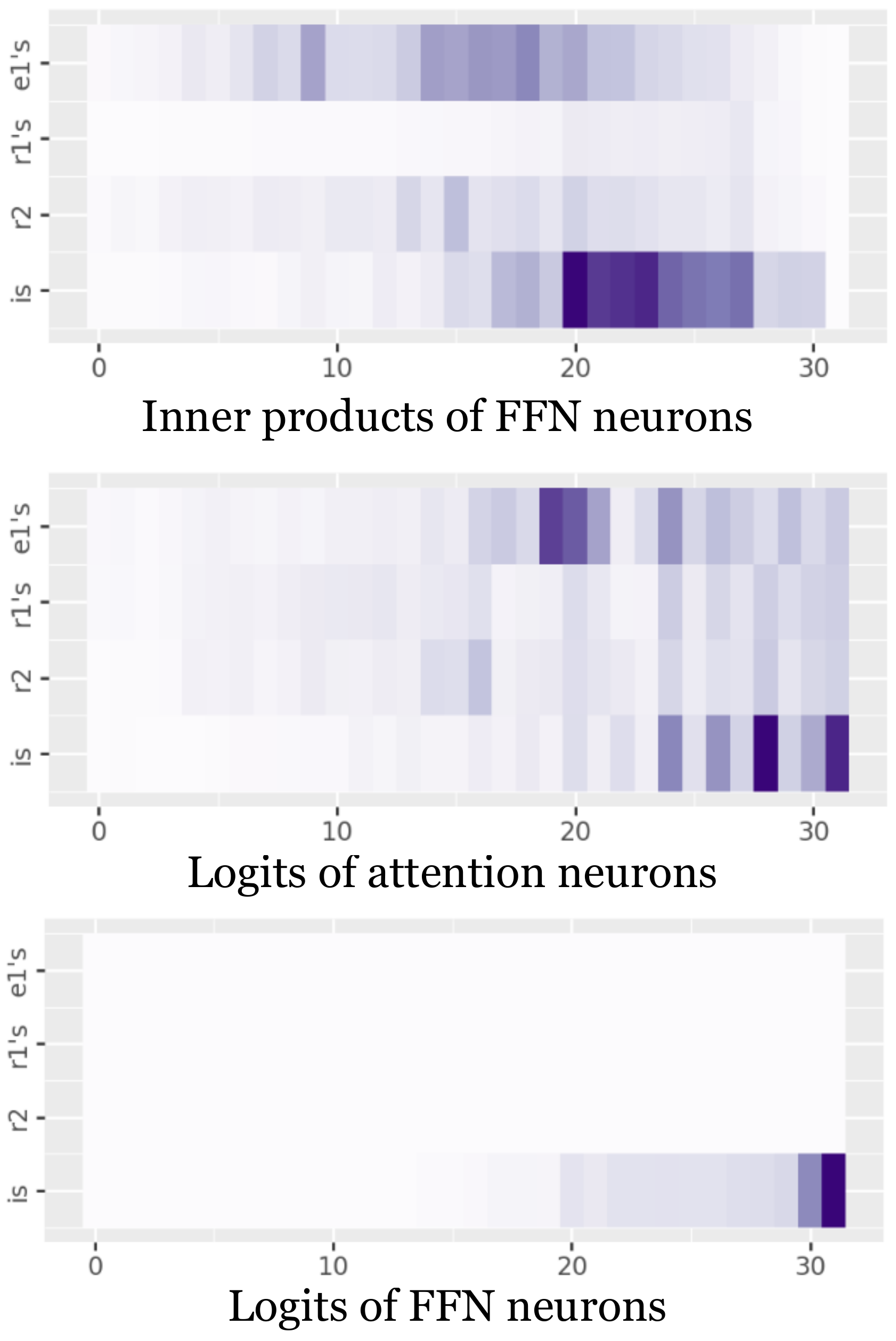}
  \caption{Results of logit flow on two-hop queries ``e1's r1's r2 is'' \texttt{->} ``e3''. The importance of relation positions (r1 and r2) is lower than single-hop queries.}
\vspace{-10pt}
\end{figure}

The results of logit flow on the two-hop queries ``e1's r1's r2 is'' \texttt{->} ``e3'' are shown in Figure 10. Compared to the logit flow results on single-hop queries (Figure 2), the importance of relation positions is significantly lower. This suggests that e1's features at the e1 position are primarily extracted into the last position, potentially activating the parameters associated with ``e1's r1'', ``e1's r2'', and ``e1's r1's r2''. This motivates our exploration between the correct and false human->human->human cases in Section 4.

\section{Results of Activation Patching on Correct and False Two-Hop Queries in Llama2-7B}

The results of activation patching on correct and false human->human->human cases in Llama2-7B are shown in Figure 11. Compared with the correct cases, the false cases show a much clearer influence at r1 position's high layers. This trend is similar to the findings of logit flow method (Figure 3), indicating that the r1 position's high features increase the probability of ``e2'', thereby reducing the accuracy of two-hop reasoning.

\begin{figure}[thb]
  \centering
  \includegraphics[width=0.88\columnwidth]{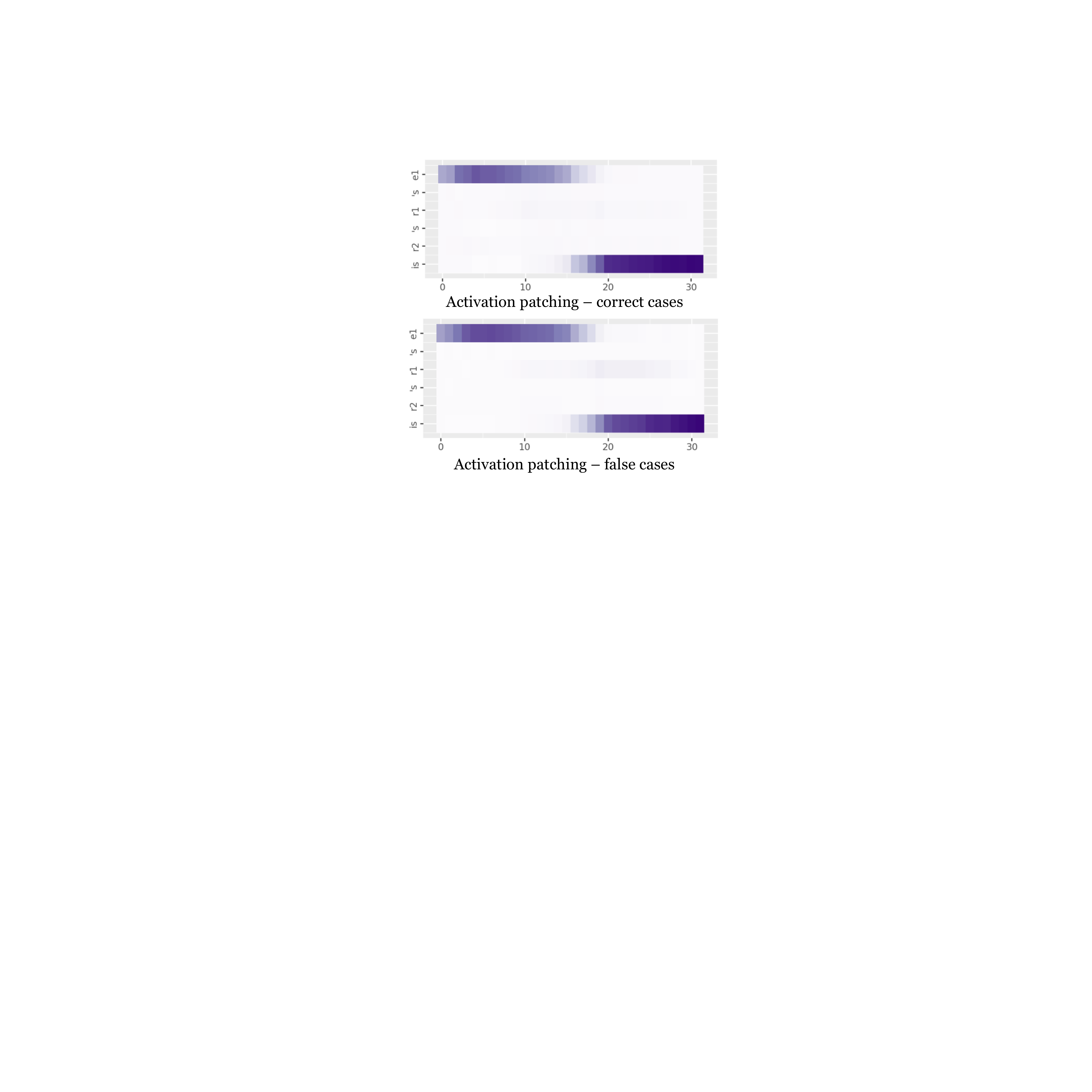}
  \caption{Results of activation patching on correct and false human->human->human cases in Llama2-7B. The importance of r1 position is 1.66\% in correct cases and 5.43\% in false cases.}
\vspace{-10pt}
\end{figure}

\section{Results of Logit Flow and Activation Patching on Correct and False Two-Hop Queries in Llama3.1-8B and Llama3.2-3B}

\begin{figure}[thb]
  \centering
  \includegraphics[width=0.88\columnwidth]{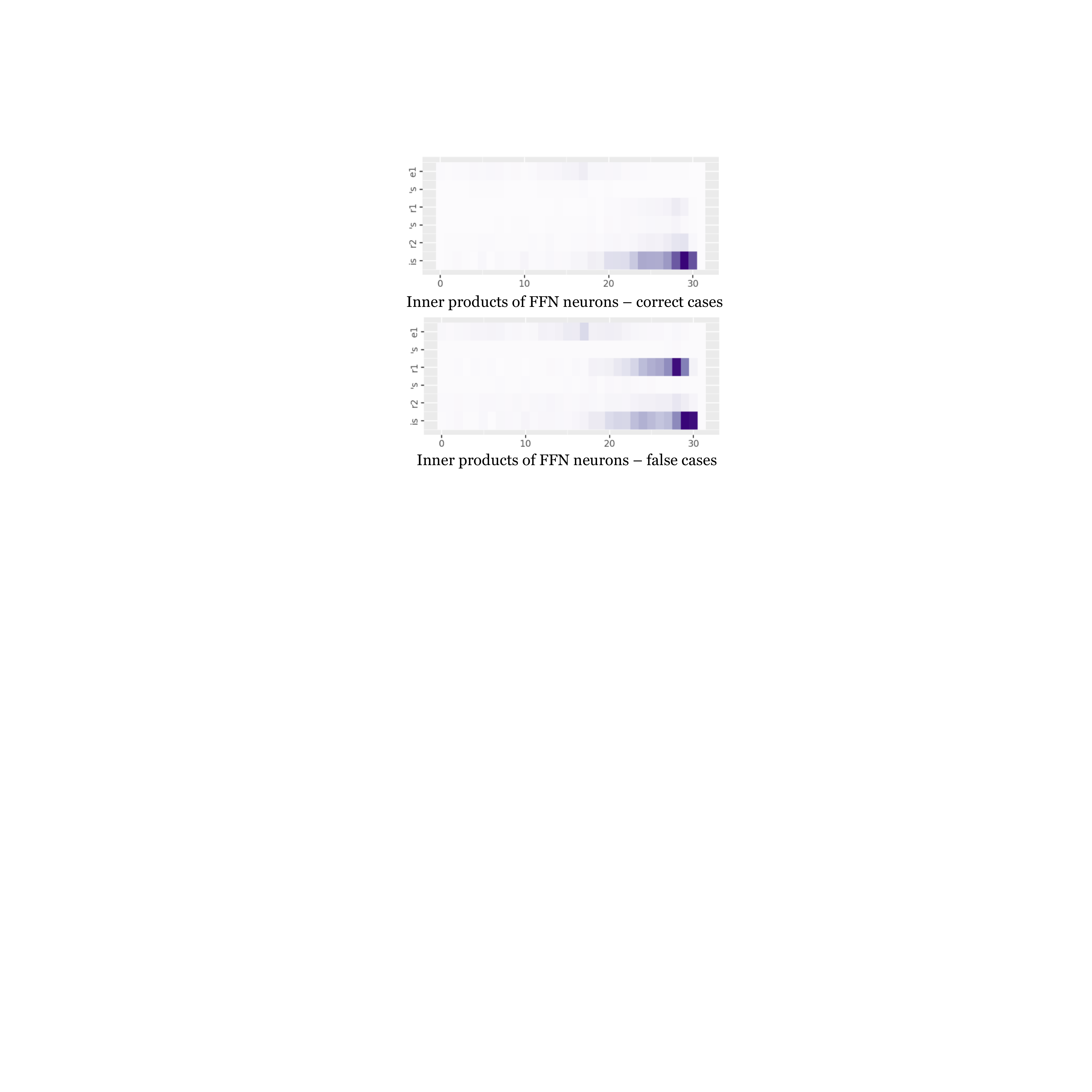}
  \caption{Results of logit flow on correct and false human->human->human cases in Llama3.1-8B. The importance of r1 position is 6.38\% in correct cases and 32.18\% in false cases.}
\vspace{-10pt}
\end{figure}

\begin{figure}[thb]
  \centering
  \includegraphics[width=0.88\columnwidth]{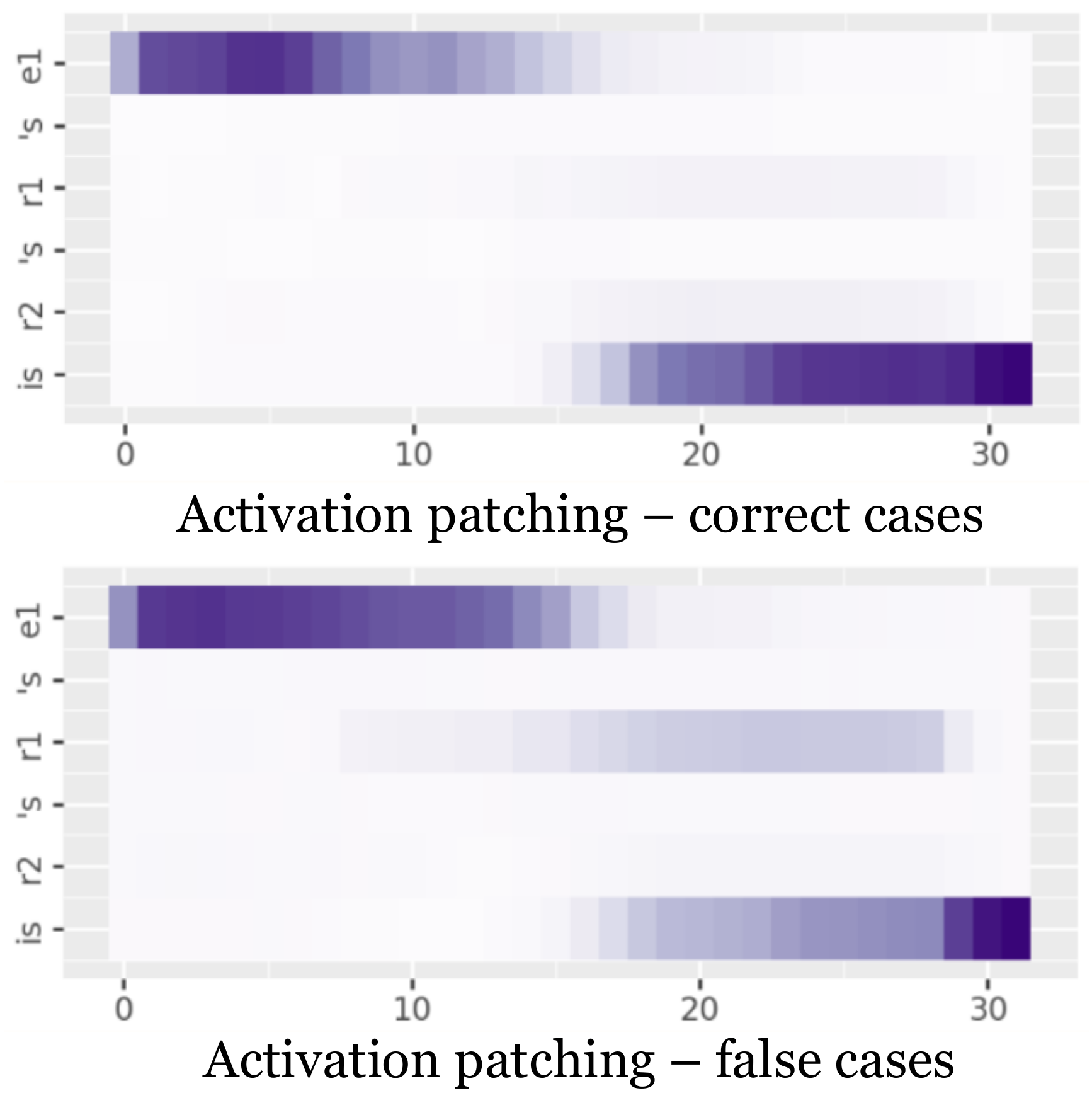}
  \caption{Results of activation patching on correct and false human->human->human cases in Llama3.1-8B. The importance of r1 position is 4.98\% in correct cases and 18.00\% in false cases.}
\vspace{-10pt}
\end{figure}

\begin{figure}[thb]
  \centering
  \includegraphics[width=0.88\columnwidth]{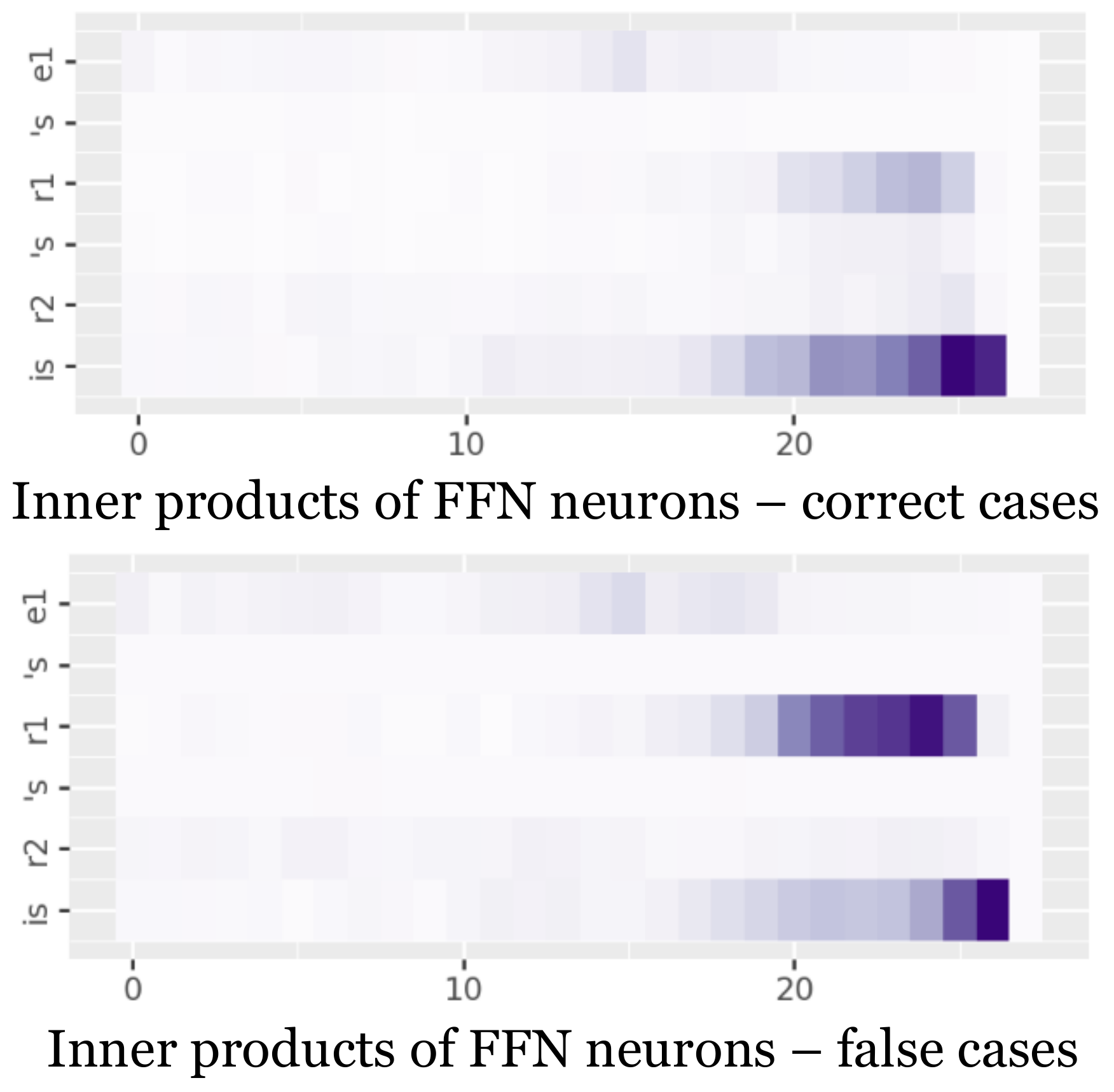}
  \caption{Results of logit flow on correct and false human->human->human cases in Llama3.2-3B. The importance of r1 position is 17.50\% in correct cases and 40.36\% in false cases.}
\vspace{-10pt}
\end{figure}

\begin{figure}[thb]
  \centering
  \includegraphics[width=0.88\columnwidth]{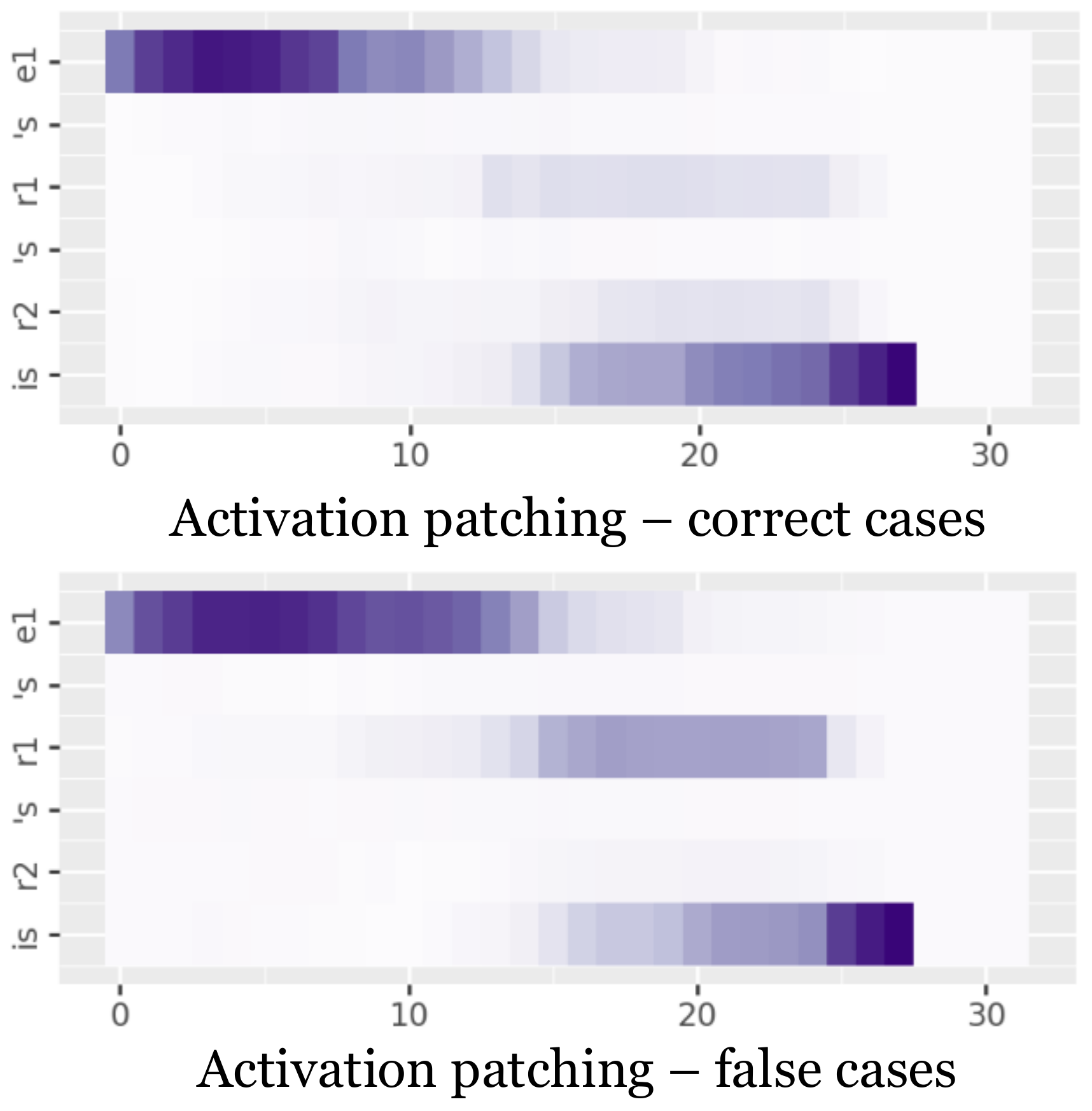}
  \caption{Results of activation patching on correct and false human->human->human cases in Llama3.2-3B. The importance of r1 position is 11.23\% in correct cases and 21.52\% in false cases.}
\vspace{-10pt}
\end{figure}

The comparison of correct and false human->human->human cases in Llama3.1-8B are shown in Figure 12 (results of logit flow) and Figure 13 (results of activation patching). Similar results of Llama3.2-3B are shown in Figure 14 (results of logit flow) and Figure 15 (results of activation patching). In both methods and models, the impact of r1 position's high layers in the false cases are larger than that in the correct cases. These results show similar trends with the results of Llama2-7B.

\section{Loss and Accuracy of Back Attention on 1-Layer Transformer}

The loss and accuracy of 1-layer transformer, 1-layer transformer with back attention, and 2-layer transformer are shown in Figure 16. The performance of 1-layer transformer with 2-layer transformer is similar, much better than that of 1-layer transformer.

\begin{figure}[thb]
  \centering
  \includegraphics[width=0.97\columnwidth]{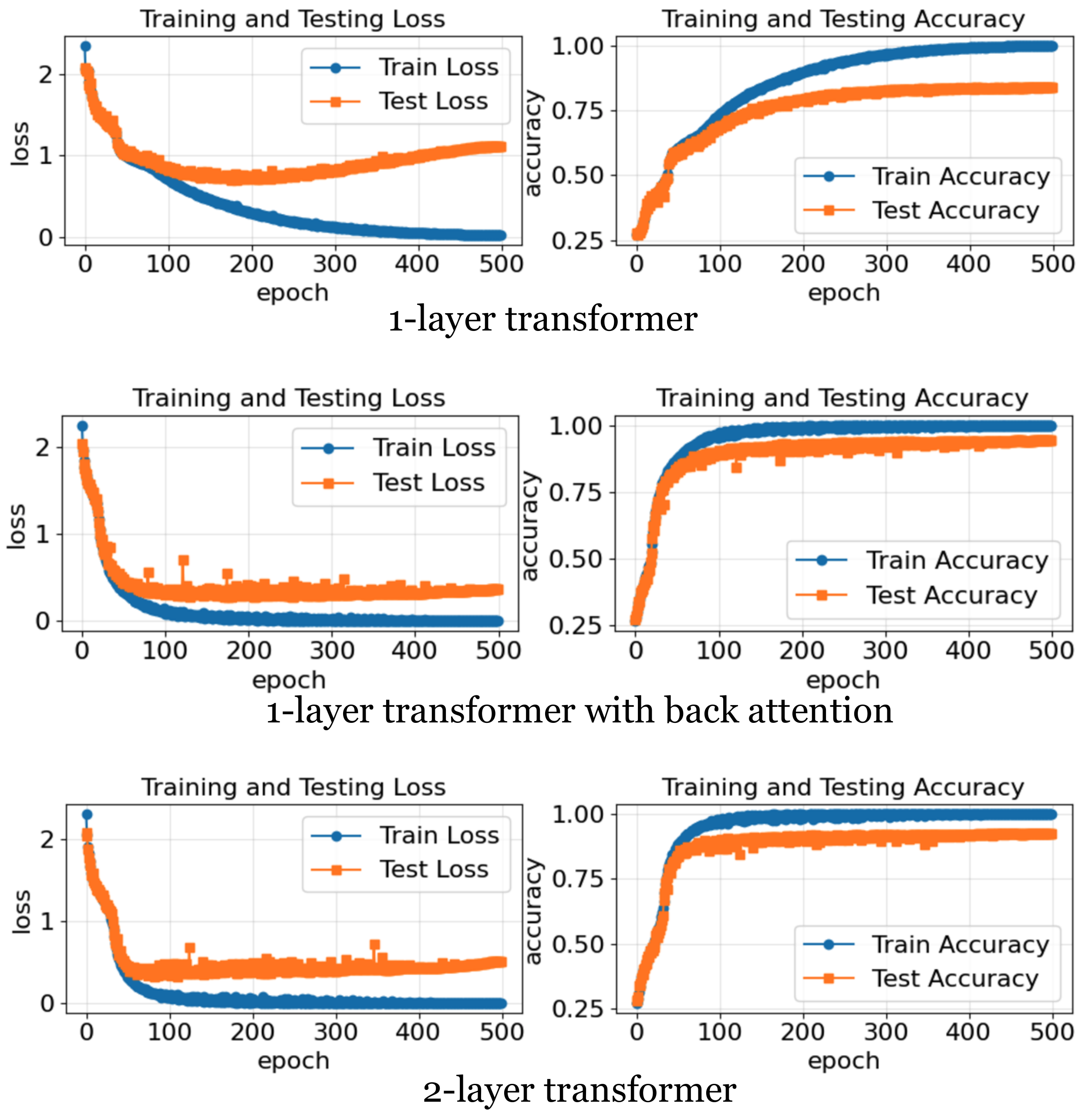}
  \caption{Loss (left) and accuracy (right) on arithmetic dataset of 1-layer transformer, 1-layer transformer with back attention, and 2-layer transformer.}
\vspace{-10pt}
\end{figure}

\end{document}